\documentclass{llncs}
\usepackage{times}
\usepackage{amssymb}
\usepackage{amstext}
\usepackage{amsmath}
\usepackage{eucal}
\usepackage{graphics}
\usepackage{graphicx}
\usepackage[T1]{fontenc}
\usepackage{ae,aecompl}
\usepackage{tikz}
\begin{document}

\pagestyle{headings}
\mainmatter

\title{Using Semantic Wikis for Structured Argument in Medical Domain}
\author{Adrian Groza\inst{1}
\and Radu Balaj\inst{1}}
\institute{Technical University of Cluj-Napoca\\
Department of Computer Science\\
Baritiu 28, RO-400391 Cluj-Napoca, Romania\\
\email{adrian@cs-gw.utcluj.ro,radu.balaj@student.utcluj.ro}\\
}

\maketitle
\thispagestyle{empty}

\abstract{
This research applies ideas from argumentation theory in the context of semantic wikis, aiming to provide support for structured-large scale argumentation between human agents.
The implemented prototype is exemplified by modelling the MMR vaccine controversy.
}

\section{Motivation}
At the moment, there is an escalate of the individuals awareness and interest with respect to the drugs they consume, possible side effects, or related symptoms, in the context of some health-related scandals such as RotaShield vaccine in 1999, GlaxoSmithKline's vaccine in 2010, or the government policies against AH1N1 in 2010.
In many cases, forums, blogs, or wikis are the first source of information when one starts searching for health services like: "best pediatrics physician in neighbourhood", "side effects of rotarix vaccine" or "the need to vaccinate against swine flu".
The main issues regard finding the $relevant$ information and $trusting$ that information when shaping ones own opinions to support justified decisions.
We approach these challenges by applying the work done in argumentation theory in the context of semantic wikis, aiming to build large scale of structured health-related argument corpus.
Our work enacts the idea of argumentative web as envisaged in~\cite{RahwanZR07} by facilitating the
semantic annotation of arguments by a large mass of users acting as a social machine~\cite{Hendler10}.

\section{Argumentation in Semantic Wiki}

\paragraph{Argument representation.}
\begin{figure}
\begin{footnotesize}
\begin{tabular}{ll}
 \hspace{0.8cm}  $A_1$ & \textit{:\ E\ asserts\ that\ A\ is\ known\ to\ be\ true.}\\
 \hspace{0.8cm}  $A_2$ & \textit{:\ E\ is\ an\ expert\ in\ domain\ D.}\\
\hspace{0.8cm}   $C$ & \textit{:\ A\ may\ (plausibly)\ be\ taken\ to\ be\ true.}\\
\hspace{0.8cm}$CQ_1$ &\textit{:\ Expertise-\ How\ credible\ is\ expert\ E\ as\ an\ expert\ source?}\\
\hspace{0.8cm}$CQ_2$ &\textit{:\ Field-\ Is\ E\ an\ expert\ in\ the\ field\ that\ the\ assertion,\ A,\ is\ in?}\\
\hspace{0.8cm}$CQ_3$ &\textit{:\ Opinion-\ Does\ E's\ testimony\ imply\ A?}\\
\hspace{0.8cm}$CQ_4$&\textit{:\ Trustworthiness-\ Is\ E\ reliable?}\\
\hspace{0.8cm}$CQ_5$&\textit{:\  Consistency-\ Is\ A\ consistent\ with\ the\ testimony\ of\ other\ experts?}\\
\hspace{0.8cm}$CQ_6$ &\textit{:\ Backup\ Evidence-\ Is\ A\ supported\ by\ evidence?}\\
\end{tabular}
\caption{Critical questions affects the credibility of the conclusion.}
\label{fig:AS_EXPERT}
\end{footnotesize}
\end{figure}

Argumentation schemes encapsulate common patterns of human reasoning such as: argument from popular opinion, argument from sign, argument from evidence, argument from position to know, or argument from expert opinion (figure~\ref{fig:AS_EXPERT}).
Argumentation schemes are defined by a set of premises $A_i$, a conclusion $C$, and a set of critical questions $CQ_i$.
When a $CQ$ is conveyed the credibility of the conclusion is decreased.
$CQs$ have the role to guide the argumentation process by providing the parties with a subset from the most encountered possible counter-arguments.

\begin{figure}
\begin{scriptsize}
\begin{tabular}{ll}
\hspace{0.75cm}$CQ_1:$ & $as\_eo(?a) \wedge hasSource(?a,?s) \wedge hasDom(?a,?d) \wedge isExpIn(?s,?d) \rightarrow o1(?a,1)$\\
\hspace{0.75cm}$CQ_2:$ & $as\_eo(?a) \wedge hasProp(?a,?p) \wedge hasDom(?a,?d) \wedge isPartOf(?s,?p)\rightarrow o2(?a,1)$\\
\hspace{0.75cm}$CQ_5:$ & $as\_eo(?a) \wedge hasProp(?a,?p) \wedge Expert(?e) \wedge hasDom(?a,?d)\wedge isExpIn(?e,?d)$\\ &$ \wedge supportsProposition(?e,?p) \rightarrow sqwrl:count(?e)$\\
& $as\_eo(?a) \wedge hasProp(?a,?p) \wedge Expert(?e) \wedge hasDom(?a,?d) \wedge isExpIn(?e,?d) $\\ 
&$\wedge nonsupportsProp(?e,?p) \rightarrow sqwrl:count(?e)$\\
\end{tabular}
\end{scriptsize}
\caption{Modelling CQs with SWRL rules.}
\label{fig:swrl}
\end{figure}

\paragraph{Argument reasoning.}
At the technical level, for the semantic annotation of arguments we use the semantic templates of the Semantic Media Wiki (SMW) framework.
The related arguments are exported from SMW in Protege, where the strength of the argument is computed in Jess based on the conveyed critical questions, represented by SWRL rules (figure~\ref{fig:swrl}).
Here $arg:o1(?a,1)$ assigns to the objection $o1$ of the argument $a$, the degree of support $1$.
In this case, the ontology is updated with the information that the appeal to authority is still valid.
After all the objections are checked, a rule computes the average mean of all objections (using mathematical built-ins such as \textit{swrlb:add}) and the data property $hasCredibility$ of an argument is set.

The available domain knowledge from the imported ontologies can also help the process of computing the strengths of the given argument.
Thus, the rule modelling $CQ_1$ is further refined if the source is not an expert in field $d_1$, but in another field $d_2$ from the same domain: $d_1 \equiv Pediatrics \sqsubseteq Medicine$ and $d_2\equiv Neurology \sqsubseteq Medicine$.
The set of all expertise fields is represented as a graph $G= \{ V, E \}$, where $V=\{v | v=field\ of\ expertise\}$ and $E=\{(u,v) | u,v=nodes \wedge v \sqsubseteq u \}$,
with the root node $LifeSciences$.
The closer to the leaf $l$ to which the subject of debate is associated with, the greater the credibility of the node representing the expert's field of expertise $e$.
Formally: $\frac{\left|{path(root,l) \cap path(root,e)}\right|}{\left|{path(root,l)}\right|}$.
This follows the principle: the larger the field, the weaker the credibility.
In order to estimate the strength of $CQ_5$, one has to count how many experts who have made a statement believe that MMR vaccine causes autism, and how many support the opposite conclusion (figure~\ref{fig:swrl}). 
Finally, the value of credibility can be assessed by dividing the number of experts who disagree with the hypothesis and the total number of experts who have made a statement about the issue.



\paragraph{Querying the argument corpus.}
The proposed framework facilitates searching based on the following criteria:
i) search by scheme: "Give only the arguments from expert opinion for supporting the argument \textit{antibiotics are not recommended for pregnant women};
ii) search by wikipedia metadata, in which specific wiki-related terms can be used to limit or refine the searching domain,
such as
1) user: "Give all the arguments of Dr. Oz user against \textit{eating meat},
2) data: "List all the arguments posted from yesterday against \textit{vaccinate against MMR},
or
3) location: "Give all the arguments of the users from Europe against \textit{genetic modified food}.
By exploiting domain knowledge like $Germany \sqsubseteq Europe$, the system is able to include in the answer the users from Germany too.

\section{Running Scenario}
\begin{figure}
    \centering
    \includegraphics[width=4.5in]{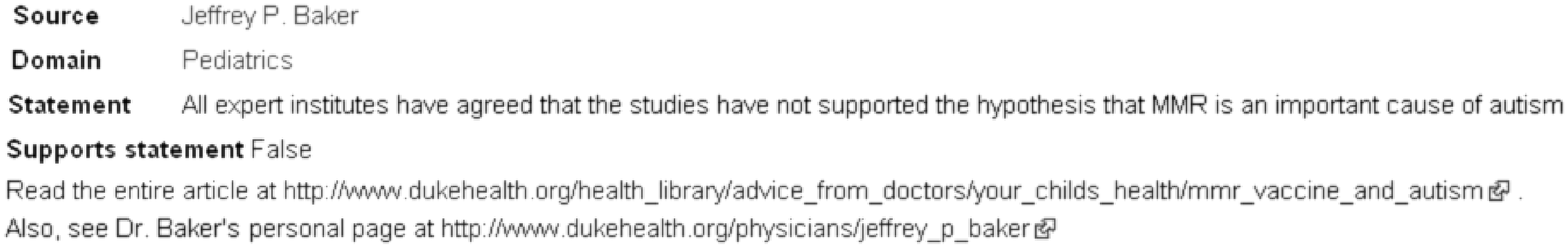}
    \caption{Attacking the hypothesis based on the expert opinion scheme.}
\label{fig:pediatru}
\end{figure}

Consider the debate regarding the topic of vaccination and whether it can cause autism in children.
The hypothesis is attacked by a pediatrician who instantiates the argument from expert opinion pattern (see figure~\ref{fig:pediatru}).
A different opinion is given by a mother who correlates the MMR vaccine with autism, by filling the template for cause to effect argumentation scheme.
When creating the arguments, the disputants can use standard terms and concepts provided by the imported ontologies in SMW, capability provided by the Semantic Gardening extension.
Here, the cause field MMR vaccine is annotated with the concept "VO\_0000731" defined in the \textit{Vaccine Ontology} (www.violinet.org) by the subsumption chain:
$VO\_0000731 \sqsubseteq VO\_0000641 \sqsubseteq VO\_0000001 \sqsubseteq OBI\_0000047 \sqsubseteq MaterialEntity \sqsubseteq IndependentContinuant \sqsubseteq Continuant \sqsubseteq Entity$.

\begin{figure}
    \centering
    \includegraphics[width=11cm]{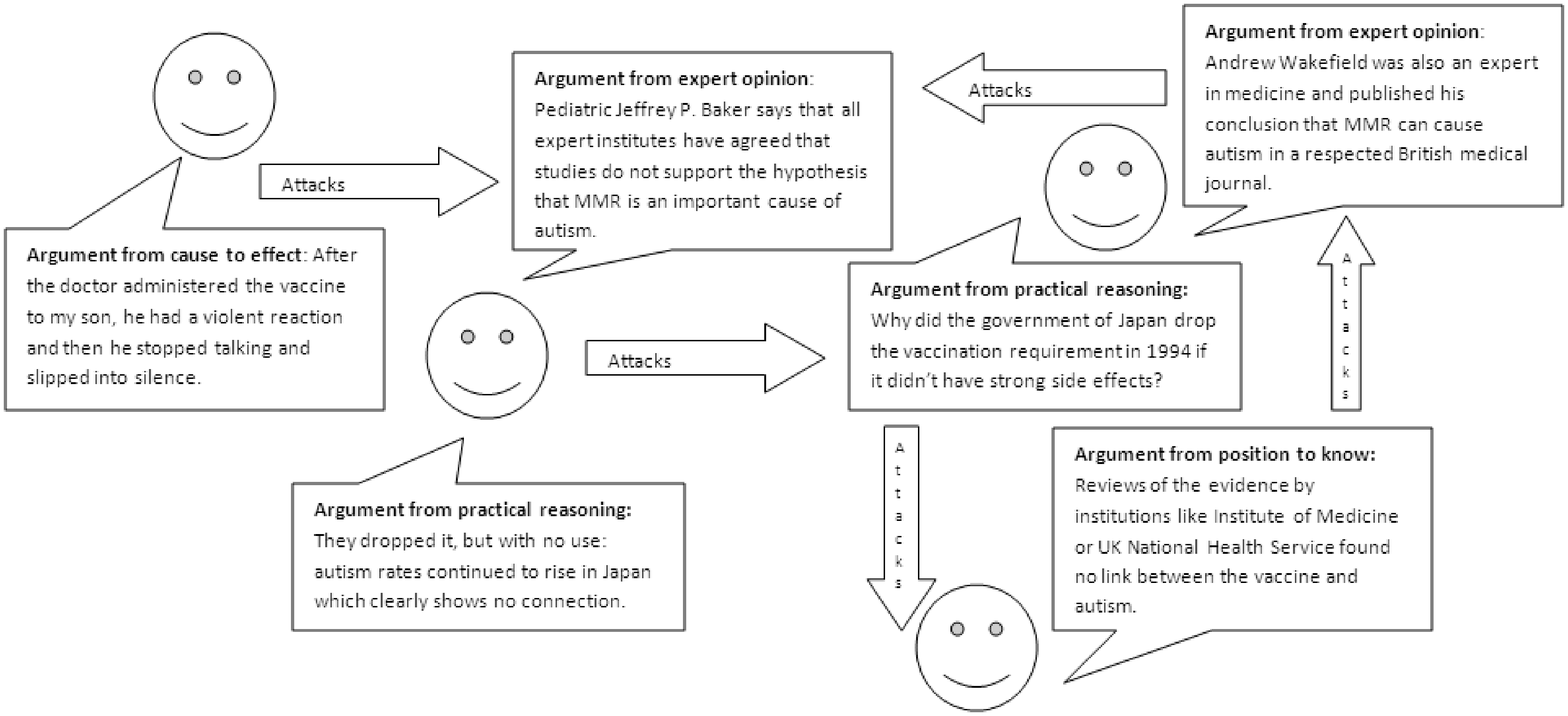}
    \caption{Argumentation scenario for MMR vaccine controversy.}
\label{fig:scenario}
\end{figure}

\begin{figure}
\begin{scriptsize}
\begin{tabular}{ll}
\hspace{0.8cm}$CQ_1:$& $AS\_EO(eo1) \wedge hasSource(eo1,Jeffrey\_P\_Baker) \wedge hasDomain(eo1,$ \\ &
$Pediatrics) \wedge isExpertIn(Jeffrey\_P\_Baker,Pediatrics) \rightarrow O1(eo1,1)$\\
\hspace{0.8cm}$CQ_2:$ & $AS\_EO(eo1) \wedge hasProp(eo1,Statement) \wedge hasDomain(eo1,Pediatrics) \wedge $\\
& $isPartOf(Statement,Pediatrics)\rightarrow O2(eo1,1)$\\
\end{tabular}
\end{scriptsize}
\caption{Instantiating SWRL rules.}
\label{fig:swrli}
\end{figure}

\section{Discussion and Conclusion}
Semantic wikis are exploited within a medical context for collaborative knowledge acquisition, annotation, and integration: WikiNeuron, WikiHit, WikiProteins, BOWiki, LexWiki, or the Hesperian Online Digital Library~\cite{Boulos09}.
Our approach is in line with~\cite{Lsbe09} which advocates the advantages of semantic wikis to exploit structured information.

Persuasive argumentation for consumer health care is analysed in~\cite{Walton10} with the help of argumentation schemes.
By enhancing drug consumers with the ability to annotate side effects might help the regulatory bodies or pharmacology industry to identify problems with newly launched drugs.
One goal is to build large scale argumentation corpora for the health care domain.
Based on hierarchical argumentation frameworks, users can navigate between medical arguments with different levels of technical specificity, in order to understand the language and the reasoning chain.

\section*{Akcnowledgements}
This work has been co-funded by the Sectorial Operational Programme Human Resources Development 2007-2013 of the Romanian Ministry of Labour, Family and Social Protection through the Financial Agreement POSDRU/89/1.5/S/62557 and PNII-Idei 170 CNCSIS.

\bibliographystyle{splncs}
\bibliography{mswa}

\end{document}